\documentclass[a4paper,fleqn]{cas-dc}
\usepackage{caption}
\usepackage{amsmath}
\usepackage{amssymb}
\usepackage{algorithm}
\usepackage{array}
\usepackage{algpseudocode}
\usepackage{bbm}
\usepackage{graphicx}
\usepackage{tabularx} 
\usepackage{booktabs} 
\usepackage[numbers]{natbib}
\usepackage{geometry}
\def\tsc#1{\csdef{#1}{\textsc{\lowercase{#1}}\xspace}}
\tsc{WGM}
\tsc{QE}
\tsc{EP}
\tsc{PMS}
\tsc{BEC}
\tsc{DE}

\begin{document}
\enlargethispage{2cm} 
\let\WriteBookmarks\relax
\def\floatpagepagefraction{1}
\def\textpagefraction{.001}

\shorttitle{LEC-KG: A LLM-Embedding Collaborative Framework}

\shortauthors{Zeng et al.}

\title[mode=title]{LEC-KG: A LLM-Embedding Collaborative Framework for Domain-Specific Knowledge Graph Construction — A Case Study on SDGs}

\affiliation[1]{organization={Computer Network Information Center, Chinese Academy of Sciences},
    city={Beijing},
    postcode={100190}, 
    country={China}}

\author[1]{Yikai Zeng}[style=chinese]
\ead{ykzeng@cnic.cn}  
\credit{Conceptualization, Methodology, Software, Writing - Original draft}

\author[1]{Yingchao Piao}[style=chinese]
\ead{pyc@cnic.cn}
\credit{Investigation, Methodology, Validation}

\author[1]{Changhua Pei}[style=chinese]
\ead{chpei@cnic.cn}
\credit{Writing - Review \& Editing, Supervision}

\author[1]{Jianhui Li}[style=chinese]
\cormark[1] 
\ead{lijh@cnic.cn}
\credit{Resources, Supervision, Project administration, Funding acquisition}

\cortext[cor1]{Corresponding author}

\begin{abstract}
Constructing domain-specific knowledge graphs from unstructured text remains challenging due to heterogeneous entity mentions, long-tail relation distributions, and the absence of standardized schemas. We present \textbf{LEC-KG}, a bidirectional collaborative framework that integrates the semantic understanding of Large Language Models (LLMs) with the structural reasoning of Knowledge Graph Embeddings (KGE). Our approach features three key components: (1) hierarchical coarse-to-fine relation extraction that mitigates long-tail bias, (2) evidence-guided Chain-of-Thought feedback that grounds structural suggestions in source text, and (3) semantic initialization that enables structural validation for unseen entities. The two modules enhance each other iteratively---KGE provides structure-aware feedback to refine LLM extractions, while validated triples progressively improve KGE representations. We evaluate LEC-KG on Chinese Sustainable Development Goal (SDG) reports, demonstrating substantial improvements over LLM baselines, particularly on low-frequency relations. Through iterative refinement, our framework reliably transforms unstructured policy text into validated knowledge graph triples.
\end{abstract}


\maketitle

\section{Introduction}

The United Nations Sustainable Development Goals (SDGs) comprise 17 interconnected objectives requiring synthesis of heterogeneous information from policy documents, governmental reports, and statistical data~\cite{UN2015agenda,sachs2019six}. Knowledge graphs (KGs) provide a structured paradigm for organizing such complex knowledge, and their integration with Large Language Models (LLMs) through Retrieval-Augmented Generation has demonstrated potential for decision support~\cite{hogan2021knowledge,pan2024unifying}. However, automatically constructing domain-specific KGs from unstructured text remains difficult, particularly in specialized domains like SDGs where standardized schemas are absent~\cite{abu2021domain,kejriwal2019domain}. These difficulties motivate our investigation into the specific challenges of SDG-oriented KG construction.

Two key challenges hinder progress in this area. \textbf{First}, the SDG domain lacks a standardized ontology for consistently categorizing entities and relations, complicating entity normalization when the same concept appears with regional or institutional variations~\cite{abu2021domain,shen2015entitylinking}. \textbf{Second}, relation distributions in real-world corpora follow a long-tail pattern: generic relations dominate while policy-relevant interactions—such as synergies and trade-offs among targets—receive limited supervision yet are central to integrated decision-making~\cite{zhang2019long,nilsson2016map}.

Existing approaches exhibit complementary limitations that prevent them from fully addressing these challenges. Traditional pipeline methods require extensive labeled data and domain-specific engineering, limiting adaptability~\cite{abu2021domain}. LLM-based approaches enable zero-shot extraction but tend to generate hallucinated content and underperform on infrequent relation types, as their pre-training corpora exhibit similar class imbalance~\cite{agrawal2024knowledge,pan2024unifying}. Knowledge Graph Embedding (KGE) methods such as RotatE~\cite{sun2019rotate} capture structural patterns effectively but cannot process unstructured text directly. Moreover, when applying KGE to evolving document collections, unseen entities pose a fundamental obstacle—embeddings cannot be computed for entities absent from training data~\cite{hamaguchi2017knowledge,shah2019open}. No existing framework bridges LLMs' semantic understanding with KGE's structural reasoning while addressing these practical constraints.

To address these limitations, we propose \textbf{LEC-KG}, a bidirectional collaborative framework. KGE provides structure-aware feedback to guide LLM extraction through evidence-grounded reasoning, while LLM-validated triples progressively improve KGE representations through uncertainty-based active selection. To handle unseen entities that emerge in new documents, we introduce a semantic initialization strategy that projects entity mentions into the KGE space via learned alignment. For the long-tail problem, we employ hierarchical relation classification that reduces the search space for infrequent relations.

The main contributions of this paper are:
\begin{itemize}
    \item A \textbf{bidirectional collaborative architecture} enabling mutual enhancement between LLMs and KGE through two channels: evidence-guided Chain-of-Thought feedback (KGE$\rightarrow$LLM) and uncertainty-based active selection (LLM$\rightarrow$KGE).
    
    \item A \textbf{semantic initialization strategy} that projects unseen entities into the KGE space via learned alignment, enabling structural validation for entities absent from training data.
    
    \item Comprehensive experiments on Chinese SDG reports demonstrating 36.8\% Micro-F1, an 11.2 point improvement over LLM few-shot extraction, with doubled performance on tail relations (13.3\% vs 6.7\%).
\end{itemize}

\section{Related Work}

\subsection{Knowledge Graph Construction from Text}

Knowledge graph construction aims to automatically extract entities, relations, and facts from unstructured text to build structured representations~\cite{ji2022survey,martinez2020information}. This fundamental task has attracted increasing attention through large language model-based approaches~\cite{pan2024unifying,zhu2024llms}.

Traditional pipeline methods decompose construction into sequential subtasks: named entity recognition, entity linking, and relation extraction~\cite{martinez2020information}. While modular, this design suffers from error propagation and relies heavily on annotated data and feature engineering, limiting adaptability to new domains~\cite{wang2017knowledge,riedel2010modeling}. End-to-end neural methods address this limitation by jointly modeling entity and relation extraction~\cite{zeng2018extracting,wei2020novel}, but still require substantial supervised data.

LLMs enable zero-shot and few-shot extraction without task-specific fine-tuning~\cite{zhu2024llms,carta2023iterative}. However, LLM-based extraction introduces challenges: hallucination of factually incorrect triples, lack of structural consistency mechanisms, and bias toward frequent relations at the expense of rare but important types~\cite{ji2023survey,agrawal2024knowledge}. These limitations motivate complementary validation mechanisms to verify LLM-generated extractions.

\subsection{Knowledge Graph Embedding and Completion}

Knowledge graph embedding methods represent entities and relations as low-dimensional vectors while preserving structural semantics, enabling link prediction for knowledge graph completion~\cite{wang2017knowledge,cao2024knowledge}.

Translation-based models interpret relations as geometric transformations between entity vectors. TransE~\cite{bordes2013translating} pioneered this direction, while subsequent models such as TransH~\cite{wang2014knowledge} and TransR~\cite{lin2015learning} addressed limitations in handling complex relation patterns. RotatE~\cite{sun2019rotate} advances the paradigm by defining relations as rotations in complex space, capturing symmetric, antisymmetric, inverse, and compositional patterns simultaneously.

Semantic matching models—including DistMult~\cite{yang2015embedding}, ComplEx~\cite{trouillon2016complex}, and ConvE~\cite{dettmers2018convolutional}—measure triple plausibility through multiplicative interactions with varying expressiveness.

Text-aware embedding methods incorporate entity descriptions or contextual information into representations. DKRL~\cite{xie2016representation} jointly learns from structure and text via CNN-based encoders, while KG-BERT~\cite{yao2019kgbert} and KEPLER~\cite{wang2021kepler} leverage pre-trained language models to encode textual descriptions alongside structural patterns. However, these methods require entity descriptions as additional input and focus on enhancing entity representations rather than extracting new relations from unstructured documents.

Beyond text awareness, embedding methods face the long-tail challenge: real-world knowledge graphs exhibit skewed distributions where important relations have few instances, causing embeddings to underperform on rare but semantically significant relations~\cite{wang2019tackling,zhang2019long}. Addressing effective extraction from documents requires not only textual understanding but also targeted support for underrepresented relations.

\subsection{Hierarchical Approaches for Relation Extraction}

Hierarchical approaches address long-tail relation distributions through coarse-to-fine classification that progressively narrows prediction scope from general categories to specific types. Han et al.~\cite{han2018hierarchical} proposed hierarchical attention for distantly supervised extraction, exploiting semantic correlations within relation hierarchies to improve performance on infrequent relations. Subsequent work incorporated knowledge graph embeddings to model relational dependencies among class labels, enabling information propagation from frequent to rare relations~\cite{takanobu2019hierarchical}. Hierarchical structures also serve as constraints for detecting noisy instances and reducing hallucination in LLM-based systems~\cite{yu2020tohre,parekh2022improving,mihindukulasooriya2023text2kgbench}.

Despite these advances, existing hierarchical methods focus predominantly on extraction and operate independently from downstream KG completion. They do not leverage graph-level structural signals to refine predictions or propagate improvements back to the extraction stage.

\subsection{Integrating Large Language Models with Knowledge Graphs}

The complementary strengths of LLMs and knowledge graphs have motivated substantial research into their unification. Pan et al.~\cite{pan2024unifying} categorize approaches into three paradigms: KG-enhanced LLMs, which incorporate structured knowledge through retrieval-augmented generation~\cite{lewis2020retrieval} or knowledge-aware pre-training~\cite{sun2020colake}; LLM-augmented KGs, which leverage language understanding for KG tasks including extraction and completion~\cite{zhu2024llms}; and Synergized LLMs+KGs, which envisions mutual enhancement through iterative refinement.

Within the synergized paradigm, recent work explores tighter coupling between LLMs and structural embeddings. KoPA~\cite{zhang2024making} projects structural embeddings into textual space for structure-aware reasoning, while KICGPT~\cite{wei2023kicgpt} retrieves relevant triples as context for LLM-based completion. Multi-agent frameworks coordinate specialized agents for extraction, verification, and refinement~\cite{fang2024llmagents}, though these primarily coordinate LLM agents rather than establishing closed-loop feedback between semantic reasoning and structural validation.

Despite these advances, a critical gap remains: existing methods operate predominantly in one direction—structural embeddings may inform LLM predictions, but LLM outputs do not refine embedding representations. The absence of closed-loop feedback between LLM-based semantic understanding and embedding-based structural reasoning limits iterative quality improvement and global consistency. Bridging this gap motivates our proposed framework.

\section{Methodology}
\label{sec:methodology}

\begin{figure*}[t]
    \centering
    \includegraphics[width=\textwidth]{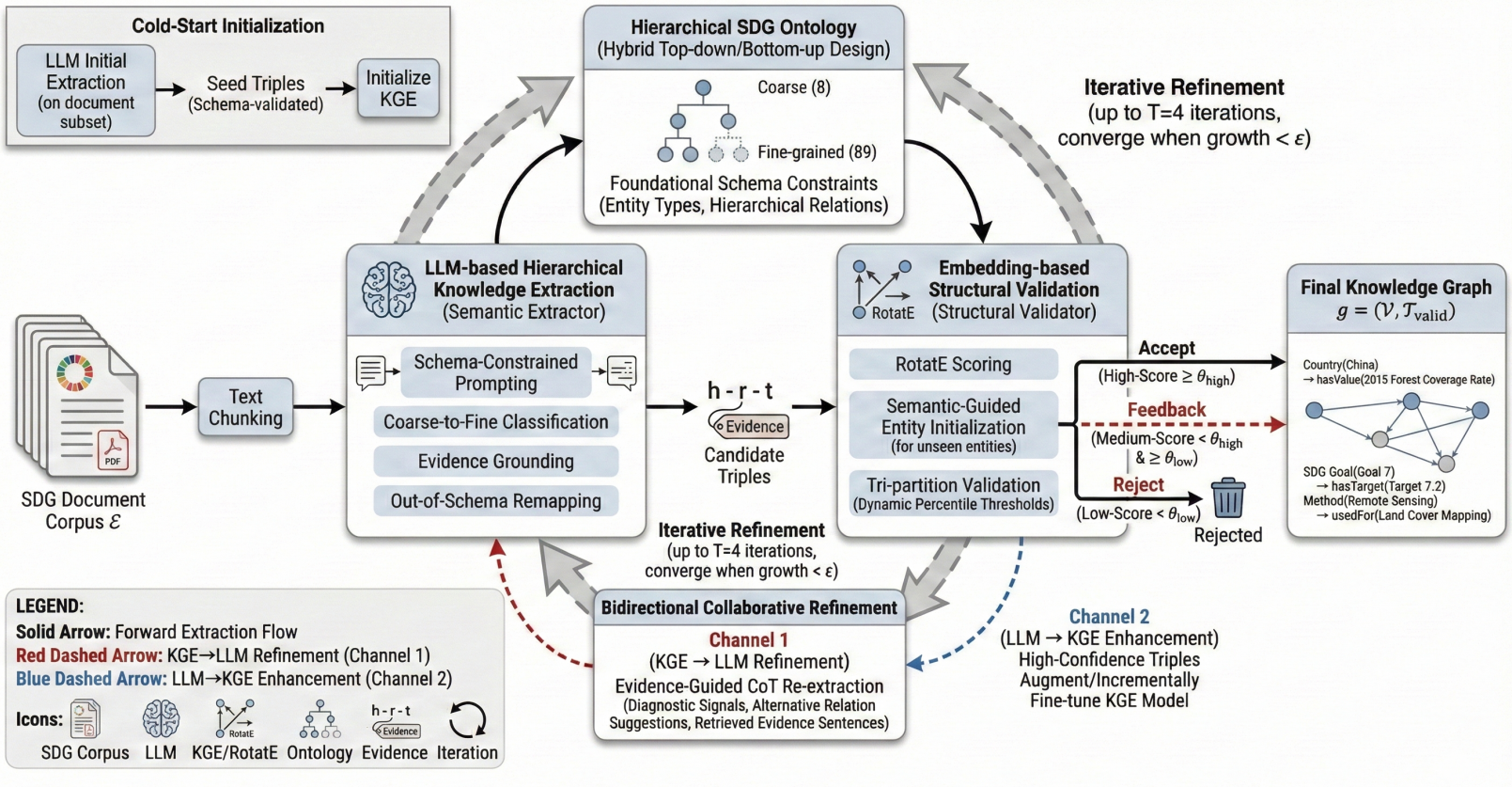} 
    \caption{Overview of the LEC-KG framework. Solid arrows indicate forward extraction flow; dashed arrows represent feedback channels: red for KGE$\to$LLM refinement and blue for LLM$\to$KGE enhancement.}
    \label{fig:framework}
\end{figure*}

\subsection{Problem Formulation and Framework Overview}
\label{sec:problem}

\subsubsection{Problem Definition}
\label{sec:problem_def}

Given a document corpus $\mathcal{D} = \{d_1, d_2, \ldots, d_n\}$ and a predefined ontology $\mathcal{O} = (\mathcal{E}, \mathcal{R}^H, \mathcal{C})$, our objective is to construct a knowledge graph $\mathcal{G} = (\mathcal{V}, \mathcal{T})$. Here, $\mathcal{E}$ denotes entity types, $\mathcal{R}^H = (\mathcal{R}^c, \mathcal{R}^f)$ represents a two-level relation hierarchy with coarse-grained categories $\mathcal{R}^c$ and fine-grained types $\mathcal{R}^f$, and $\mathcal{C}$ specifies domain constraints including permissible entity type pairs for each relation.

A triple $\tau = (h, r, t)$ is considered valid if it satisfies both schema constraints (entity types conform to $\mathcal{C}$) and structural plausibility (embedding score $s(\tau) \geq \theta$).

\subsubsection{Bidirectional Collaboration Mechanism}
\label{sec:bidirectional}

Our framework establishes bidirectional information flow between an LLM (semantic extractor) and a KGE model (structural validator), as illustrated in Figure~\ref{fig:framework}.

\paragraph{Forward Pass.}
The LLM extracts candidate triples from documents under ontology constraints (Section~\ref{sec:extraction}), with relations predicted via hierarchical classification implicitly encoded in the prompt (Section~\ref{sec:prompt}). The KGE model computes structural plausibility scores for validation (Section~\ref{sec:embedding}).

\paragraph{Backward Pass.}
Based on validation outcomes, triples are routed through two feedback channels:

\textbf{Channel 1 (KGE $\to$ LLM):} Low-scoring triples ($s(\tau) < \theta$) trigger re-extraction. The KGE model provides diagnostic feedback including alternative relation candidates, guiding the LLM to reconsider predictions with evidence from source text.

\textbf{Channel 2 (LLM $\to$ KGE):} High-confidence triples ($s(\tau) \geq \theta_{\text{high}}$, where $\theta_{\text{high}} > \theta$) augment the KGE training set, enabling progressive expansion of domain-specific pattern coverage.

The detailed feedback mechanism is presented in Section~\ref{sec:collaboration}. Iteration continues until relative growth in validated triples falls below $\epsilon$ or maximum iterations $T$ is reached.

\subsubsection{Cold-Start Initialization}
\label{sec:coldstart}

A key challenge is initializing the KGE model without pre-existing labeled triples. We adopt a two-stage cold-start strategy~\cite{shi2018open,hamaguchi2017knowledge}: (1) the LLM performs initial extraction on a document subset, producing candidate triples; (2) candidates passing schema validation in $\mathcal{C}$ serve as seed triples for KGE initialization. This bootstrapping enables the framework to operate without manual annotation while providing sufficient structural signal for subsequent validation. The semantic initialization strategy for handling unseen entities during this process is detailed in Section~\ref{sec:embedding}.

\begin{algorithm}[t]
\caption{LEC-KG: Bidirectional Collaborative KG Construction}
\label{alg:collaborative}
\begin{algorithmic}[1]
\Require Corpus $\mathcal{D}$, Ontology $\mathcal{O}$, LLM $\mathcal{M}_{\text{LLM}}$, KGE model $\mathcal{M}_{\text{KGE}}$
\Ensure Knowledge Graph $\mathcal{G} = (\mathcal{V}, \mathcal{T}_{\text{valid}})$
\Statex \textbf{Params:} Thresholds $\theta < \theta_{\text{high}}$, max iterations $T$, convergence $\epsilon$

\Statex
\Statex \textit{// Phase 1: Cold-Start Initialization}
\State $\mathcal{T}_{\text{init}} \leftarrow$ LLM extraction on subset of $\mathcal{D}$
\State $\mathcal{T}_{\text{seed}} \leftarrow \{\tau \in \mathcal{T}_{\text{init}} \mid \text{schema}(\tau) = \texttt{true}\}$
\State Initialize $\mathcal{M}_{\text{KGE}}$ on $\mathcal{T}_{\text{seed|}|}$; \; $\mathcal{T}_{\text{valid}} \leftarrow \emptyset$

\Statex
\Statex \textit{// Phase 2: Iterative Refinement}
\State $\mathcal{T}_{\text{cand}} \leftarrow$ LLM extraction on full $\mathcal{D}$
\For{$t = 1$ to $T$}
    \State Score all $\tau \in \mathcal{T}_{\text{cand|}|}$; \; accept $\{\tau \mid s(\tau) \geq \theta\}$ into $\mathcal{T}_{\text{valid}}$
    \If{relative growth $< \epsilon$} \textbf{break} \EndIf
    \State \textit{Channel 1:} Re-extract $\{\tau \mid s(\tau) < \theta\}$ with KGE feedback $\to \mathcal{T}_{\text{cand}}$
    \State \textit{Channel 2:} Fine-tune $\mathcal{M}_{\text{KGE}}$ on $\{\tau \mid s(\tau) \geq \theta_{\text{high}}\}$
\EndFor

\State \Return $\mathcal{G} = (\mathcal{V}, \mathcal{T}_{\text{valid}})$
\end{algorithmic}
\end{algorithm}

\subsection{Hierarchical SDG Ontology Design}
\label{sec:ontology}

The quality of schema-constrained extraction critically depends on a well-designed ontology. Our hierarchical SDG ontology defines entity types, relation types, and constraints following a hybrid methodology combining top-down guidance with bottom-up corpus-driven refinement.

\subsubsection{Design Methodology}
\label{sec:design_method}

Preliminary experiments with unconstrained LLM extraction from SDG documents yielded over 200 distinct relation expressions. The LLM generated semantically plausible but inconsistent expressions (e.g., ``locatedIn'', ``situatedAt'', ``withinRegionOf'' for identical spatial relationships), preventing reliable mapping to a coherent graph structure. This observation motivated a schema-constrained approach requiring systematic ontology design.

We adopted a hybrid methodology to balance domain coverage with corpus-specific patterns:

\paragraph{Top-down Guidance.}
We reference SustainGraph~\cite{fotopoulou2022sustaingraph} and the UN SDG indicator framework for foundational structure, establishing principled entity types (Goals, Targets, Indicators) and their relationships with geographical and temporal dimensions.

\paragraph{Bottom-up Refinement.}
To capture patterns specific to Chinese SDG reports, we performed iterative refinement on the 200+ extracted expressions: LLM-assisted semantic clustering followed by manual verification consolidated these into 89 standardized relation types grouped into 8 coarse-grained categories.

\begin{figure*}[t]
    \centering
    \includegraphics[width=0.8\textwidth]{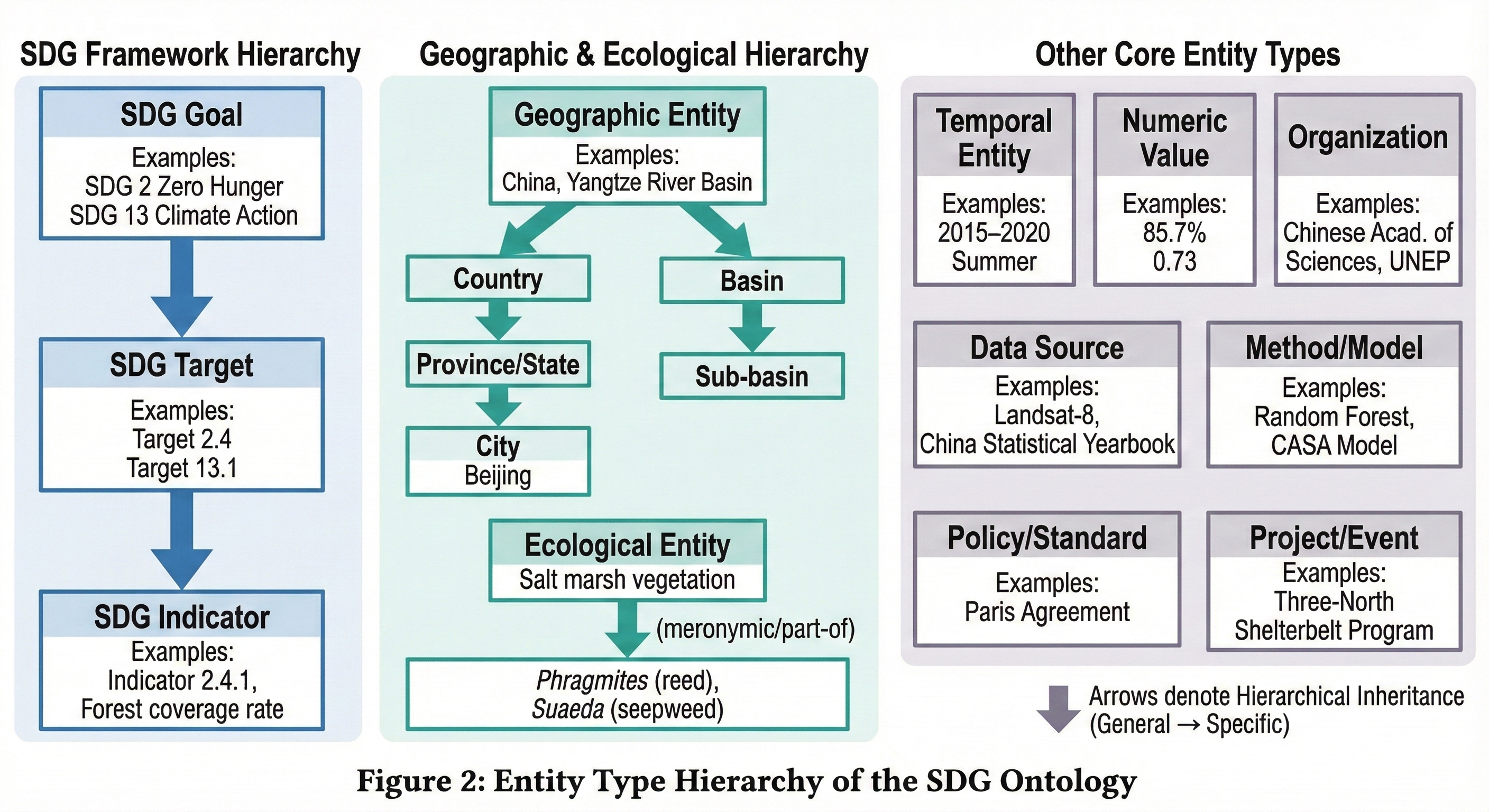}
    \caption{Entity type hierarchy. Solid arrows indicate subtype relationships; dashed arrows indicate meronymic (part-of) relationships.}
    \label{fig:entity_hierarchy}
\end{figure*}

\subsubsection{Entity Type Hierarchy}
\label{sec:entity_types}

Our ontology defines 12 entity types organized hierarchically (Figure~\ref{fig:entity_hierarchy}), capturing three taxonomic relationships:

\begin{itemize}
    \item \textbf{Administrative}: Country $\to$ Province $\to$ City; Basin $\to$ Sub-basin
    \item \textbf{SDG Framework}: Goal $\to$ Target $\to$ Indicator
    \item \textbf{Meronymic}: e.g., ``Salt marsh vegetation'' includes ``\textit{Phragmites}'' and ``\textit{Suaeda}''
\end{itemize}

Table~\ref{sec:entity_types} summarizes entity types with examples. Types are mutually exclusive at leaf level while allowing hierarchical inheritance—a relation can constrain its domain to \texttt{GeographicEntity} to accept all geographic subtypes.

\begin{table}[t]
\centering
\caption{Coarse-grained relation categories with representative fine-grained types.}
\label{tab:relation_categories}
\small
\begin{tabularx}{\columnwidth}{p{2.8cm} >{\raggedright\arraybackslash}X c}
\toprule
\textbf{Category} & \textbf{Representative Relations} & \textbf{\#} \\
\midrule
Definition \& Naming & definedAs, fullNameOf, aliasOf & 4 \\
Hierarchy \& Composition & belongsTo, contains, partOf & 4 \\
Spatiotemporal & locatedIn, timeRangeOf, covers & 11 \\
Quantitative & hasValue, hasUnit, meanValueOf & 14 \\
Trend \& Change & trendOf, changeRateOf, growthRateOf & 6 \\
Provenance \& Method & dataSourceOf, usesMethod, citedFrom & 17 \\
Causality \& Impact & causes, affects, promotes, mitigates & 10 \\
Application \& Status & usedFor, statusOf, contributesToSDG & 23 \\
\bottomrule
\end{tabularx}
\end{table}

\subsubsection{Hierarchical Relation Schema}
\label{sec:relation_schema}

\begin{figure*}[t]
    \centering
    \includegraphics[width=0.8\textwidth]{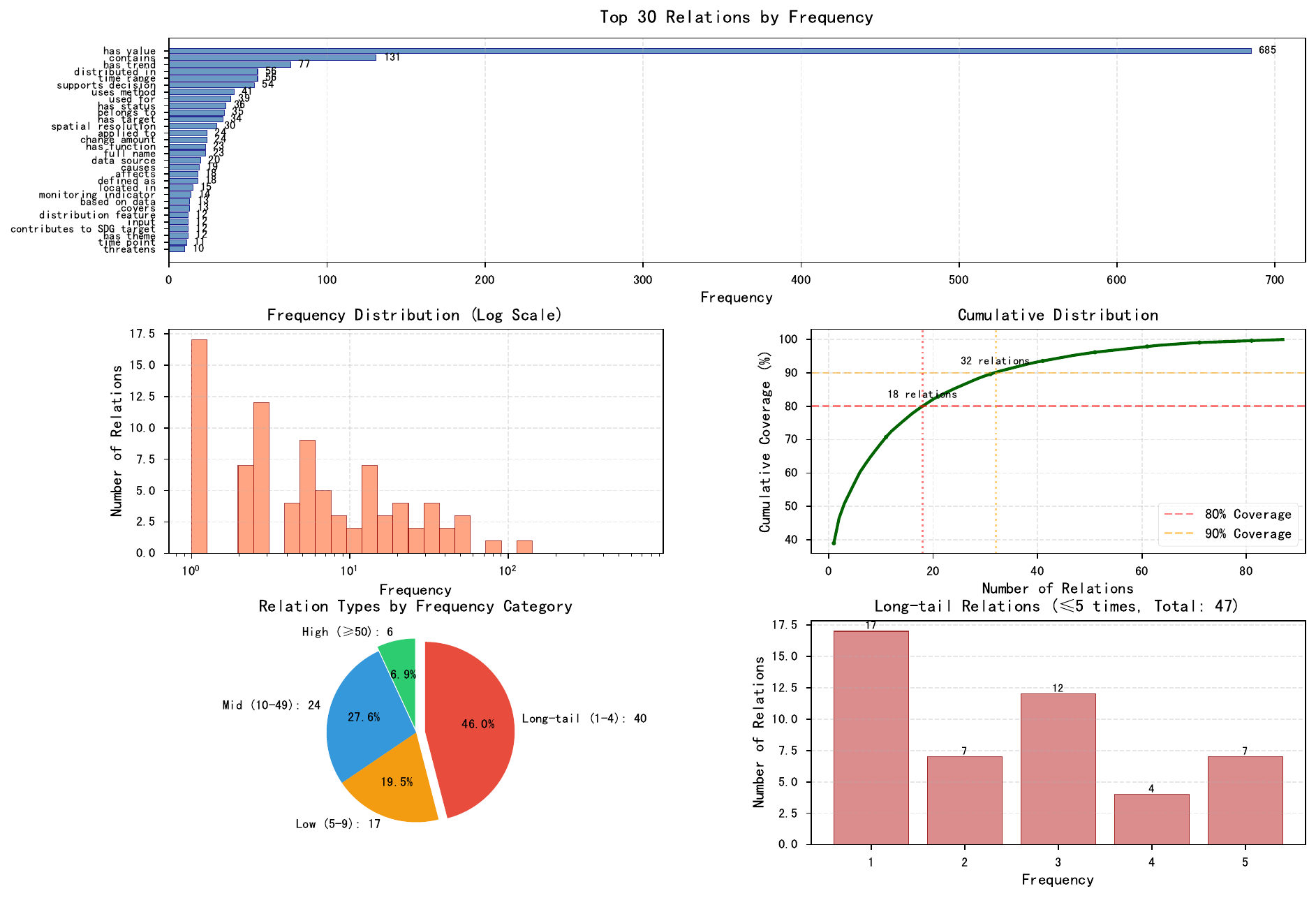}
    \caption{Long-tail distribution of relation types in our annotated dataset (1,758 triples). The top 10 relations account for 68.2\% of instances.}
    \label{fig:relation_distribution}
\end{figure*}

Our schema comprises 8 coarse-grained categories containing 89 fine-grained relation types, directly supporting the hierarchical classification strategy (Section~\ref{sec:extraction}). Statistics are computed from our manually annotated evaluation set of 1,758 triples (detailed in Section~\ref{sec:dataset}).

\paragraph{Long-tail Distribution.}
Extracted triples exhibit severe class imbalance (Figure~\ref{fig:relation_distribution}): the most frequent relation (``hasValue'') accounts for 685 instances while 23 relations appear fewer than 5 times. This long-tail distribution motivates our hierarchical approach—first classifying into coarse categories with sufficient training signal before fine-grained prediction.

\paragraph{Coarse-grained Categories.}
Table~\ref{tab:relation_categories} summarizes the 8 categories with representative relations. The complete schema with all 89 fine-grained relations is provided in Appendix~\ref{app:relation_schema}.

\begin{table}[t]
\centering
\caption{Coarse-grained relation categories with representative fine-grained types.}
\label{tab:relation_categories}
\small
\begin{tabularx}{\columnwidth}{p{2.5cm} X c}
\toprule
\textbf{Category} & \textbf{Representative Relations} & \textbf{\#} \\
\midrule
Definition \& Naming & definedAs, fullNameOf, aliasOf & 4 \\
Hierarchy \& Composition & belongsTo, contains, partOf & 4 \\
Spatiotemporal & locatedIn, timeRangeOf, covers & 11 \\
Quantitative & hasValue, hasUnit, meanValueOf & 14 \\
Trend \& Change & trendOf, changeRateOf, growthRateOf & 6 \\
Provenance \& Method & dataSourceOf, usesMethod, citedFrom & 17 \\
Causality \& Impact & causes, affects, promotes, mitigates & 10 \\
Application \& Status & usedFor, statusOf, contributesToSDG & 23 \\
\bottomrule
\end{tabularx}
\end{table}

\paragraph{Schema Constraints.}
The complete schema—including entity type definitions, hierarchical relations, and domain/range constraints—is encoded into extraction prompts to constrain the LLM's output space (detailed in Section~\ref{sec:extraction}). This prevents semantic drift where unconstrained generation produces diverse synonymous expressions for identical relationships.

\paragraph{Comparison with SustainGraph.}
Our ontology extends SustainGraph~\cite{fotopoulou2022sustaingraph} with three enhancements for Chinese scientific reports: (1) detailed provenance chains capturing data sources and processing pipelines, (2) fine-grained quantitative relations reflecting statistical content, and (3) causality relations capturing analytical findings.
\subsection{Hierarchical Knowledge Extraction}
\label{sec:extraction}

SDG relation types exhibit severe long-tail distribution where head classes dominate training signal. Direct classification without structural guidance suffers from head-class bias and schema drift. We present a hierarchical extraction strategy that leverages the coarse-to-fine ontology structure (Section~\ref{sec:ontology}) within a single-pass LLM framework using DeepSeek~\cite{deepseekai2024deepseekv3}.

\subsubsection{Single-Pass Hierarchical Extraction}
\label{sec:single_pass}

Unlike pipeline approaches that separate entity recognition from relation extraction, we adopt joint extraction where the LLM simultaneously identifies entities, relations, and supporting evidence in a single inference pass. This design reduces error propagation and leverages the LLM's ability to capture contextual dependencies.

\paragraph{Document Chunking.}
Documents are segmented into 2,000-character chunks with 200-character overlap to handle long documents while preserving local context for relation extraction.

\paragraph{Hierarchical Output Format.}
For each chunk, the LLM extracts tuples containing five components:
\begin{equation}
\tau_{\text{raw}} = (h, r, t, e, c)
\label{eq:extraction_output}
\end{equation}
where $h$ and $t$ are head and tail entities, $r$ is the fine-grained relation, $e$ is supporting evidence (verbatim text span), and $c \in \mathcal{R}^c$ is the coarse-grained relation category. The evidence $e$ grounds extractions in source text and enables evidence-guided feedback during collaborative refinement (Section~\ref{sec:collaboration}).

\paragraph{Coarse-to-Fine Classification via Prompting.}
The prompt encodes our two-level relation hierarchy (Section~\ref{sec:ontology}), requiring the LLM to: (1) first determine the coarse category $c$ from 8 options, constraining the semantic scope; (2) then select the fine-grained relation $r$ from the subset $\mathcal{R}^f(c)$ associated with that category. This reduces the effective search space from 89 relations to an average of 11 per category, mitigating long-tail bias by ensuring tail relations compete only within their semantic neighborhood.

Algorithm~\ref{alg:extraction} presents the complete extraction procedure including out-of-schema handling.

\begin{algorithm}[t]
\caption{Hierarchical Knowledge Extraction}
\label{alg:extraction}
\begin{algorithmic}[1]
\Require Document $d$, Ontology $\mathcal{O} = (\mathcal{E}, \mathcal{R}^H, \mathcal{C})$, LLM $\mathcal{M}_{\text{LLM}}$
\Ensure Candidate triples $\mathcal{T}_{\text{cand}}$

\Statex
\Statex \textit{// Phase 1: Chunking and Hierarchical Extraction}
\State $\mathcal{C}_d \leftarrow \text{ChunkDocument}(d, \text{size}{=}2000, \text{overlap}{=}200)$
\State $\mathcal{T}_{\text{raw}} \leftarrow \emptyset$
\ForAll{chunk $c \in \mathcal{C}_d$}
    \State $\mathcal{P} \leftarrow \text{BuildHierarchicalPrompt}(c, \mathcal{O|}|)$ \Comment{Coarse-to-fine}
    \State $\mathcal{T}_c \leftarrow \mathcal{M}_{\text{LLM|}|}(\mathcal{P})$ \Comment{Output: $(h, r, t, e, c)$}
    \State $\mathcal{T}_{\text{raw}} \leftarrow \mathcal{T}_{\text{raw}} \cup \mathcal{T}_c$
\EndFor
\Statex
\Statex \textit{// Phase 2: Schema Validation and Remapping}
\State $\mathcal{T}_{\text{cand}} \leftarrow \emptyset$; \; $\mathcal{T}_{\text{oos}} \leftarrow \emptyset$
\ForAll{$\tau = (h, r, t, e, c) \in \mathcal{T}_{\text{raw}}$}
    \If{$r \in \mathcal{R}^f(c)$}
        \State $\mathcal{T}_{\text{cand}} \leftarrow \mathcal{T}_{\text{cand}} \cup \{(h, r, t, e)\}$
    \Else
        \State $\mathcal{T}_{\text{oos}} \leftarrow \mathcal{T}_{\text{oos}} \cup \{\tau\}$ \Comment{Out-of-schema}
    \EndIf
\EndFor

\Statex
\Statex \textit{// Phase 3: Out-of-Schema Remapping}
\ForAll{$\tau = (h, r, t, e, c) \in \mathcal{T}_{\text{oos}}$}
    \State $r' \leftarrow \mathcal{M}_{\text{LLM}}(\text{RemapPrompt}(\tau, \mathcal{R}^f(c)))$
    \If{$r' \in \mathcal{R}^f(c)$}
        \State $\mathcal{T}_{\text{cand}} \leftarrow \mathcal{T}_{\text{cand}} \cup \{(h, r', t, e)\}$
    \EndIf
\EndFor

\State \Return $\mathcal{T}_{\text{cand}}$
\end{algorithmic}
\end{algorithm}

\subsubsection{Schema-Constrained Prompt Design}
\label{sec:prompt}

Effective prompt design is critical for schema compliance and extraction quality.

\paragraph{Ontology Encoding.}
The prompt embeds the complete hierarchical schema (Appendix~\ref{app:relation_schema}), including: entity type definitions with examples, the 8 coarse-grained categories with semantic descriptions, and fine-grained relations organized under each category.

\paragraph{Hierarchical Instruction.}
The prompt explicitly instructs coarse-to-fine reasoning:
\begin{quote}
\small\textit{``For each relation, first determine which of the 8 categories it belongs to, then select the most appropriate fine-grained relation from that category. Output both the category and the specific relation.''}
\end{quote}

\paragraph{Semantic Mapping Guidelines.}
To handle linguistic variation, the prompt includes mapping guidelines for common paraphrases (e.g., ``sourced from'' $\to$ \texttt{dataSourceOf}, ``located at'' $\to$ \texttt{locatedIn}) and directionality requirements for correct subject-object ordering.

\paragraph{Few-shot Demonstrations.}
We include 3--5 representative examples covering diverse relation categories (Table~\ref{tab:fewshot}), demonstrating the expected output format with both coarse category and fine-grained relation.

\begin{table}[t]
\centering
\caption{Few-shot demonstrations for hierarchical extraction.}
\label{tab:fewshot}
\small
\begin{tabular}{p{2.8cm}p{1.8cm}p{2.6cm}}
\toprule
\textbf{Evidence} & \textbf{Category} & \textbf{Triple} \\
\midrule
Forest coverage reached 23.04\%. & Quantitative & (Forest coverage, hasValue, 23.04\%) \\
\midrule
Data derived from MODIS satellite. & Provenance & (Data, dataSourceOf, MODIS) \\
\midrule
Study area in Yangtze River Basin. & Spatiotemporal & (Study area, locatedIn, Yangtze River Basin) \\
\midrule
Random Forest used for classification. & Provenance & (Classification, usesMethod, Random Forest) \\
\midrule
Climate change exacerbated drought. & Causality & (Climate change, exacerbates, Drought risk) \\
\bottomrule
\end{tabular}
\end{table}

\paragraph{Confidence Estimation.}
Each triple includes an LLM self-assessed confidence $p_{\text{llm}} \in [0,1]$ based on relation explicitness in the source text. Triples with $p_{\text{llm}} < 0.5$ are flagged for priority review during collaborative refinement Section~\ref{sec:collaboration}.

\subsubsection{Out-of-Schema Relation Remapping}
\label{sec:remapping}

Despite schema constraints, LLMs occasionally generate relations outside the predefined set due to semantic paraphrasing. We address this through targeted remapping.

\paragraph{Detection.}
For each extracted tuple $(h, r, t, e, c)$, we verify whether $r \in \mathcal{R}^f(c)$. Tuples failing this check are flagged as out-of-schema (OOS).

\paragraph{Evidence-Guided Remapping.}
OOS tuples are processed by a second LLM call receiving: (1) the original triple with evidence $e$; (2) the valid relation set $\mathcal{R}^f(c)$ for the predicted category; (3) instruction to select the closest semantic match or indicate no suitable match exists.

\paragraph{Fallback.}
If remapping fails, the tuple is discarded. This conservative strategy prioritizes precision over recall, as structurally invalid triples would degrade downstream KGE training.

\subsubsection{Hallucination Mitigation}
\label{sec:hallucination}

Our framework addresses hallucination through three complementary mechanisms:

\begin{itemize}
    \item \textbf{Schema-level constraint}: Relations must be selected from the predefined ontology, preventing free-form generation of inconsistent predicates.
    
    \item \textbf{Evidence grounding}: Each triple includes supporting text span $e$, enabling verification that extractions are anchored in source content.
    
    \item \textbf{Structural validation}: The KGE model (Section~\ref{sec:embedding}) assigns low plausibility scores to locally coherent but globally inconsistent triples, providing complementary validation beyond textual grounding.
\end{itemize}

\subsection{Embedding-based Structural Validation}
\label{sec:embedding}

This section details the structural validation mechanism that assesses triple plausibility from a graph-structural perspective, complementing the linguistic extraction (Section~\ref{sec:extraction}) and schema compliance ensured by the ontology (Section~\ref{sec:ontology}).

\subsubsection{RotatE for Structural Scoring}
\label{sec:rotate}

We adopt RotatE~\cite{sun2019rotate} for structural validation due to its capacity to model symmetric, antisymmetric, inversion, and composition relations prevalent in SDG domains. RotatE interprets relations as rotations in complex vector space:
\begin{equation}
s(\tau) = \sigma\left(-\|\mathbf{h} \circ \mathbf{r} - \mathbf{t}\|\right)
\label{eq:rotate}
\end{equation}
where $\mathbf{h}, \mathbf{r}, \mathbf{t} \in \mathbb{C}^d$ are complex embeddings of head entity, relation, and tail entity respectively, $\circ$ denotes element-wise product, and $\sigma$ is the sigmoid function. Higher scores indicate greater structural plausibility.

We employ 512-dimensional embeddings with self-adversarial negative sampling. Training follows an incremental strategy: cold-start initialization with extracted triples, followed by updates incorporating high-confidence validated triples at each iteration.

\subsubsection{Semantic-Guided Entity Initialization}
\label{sec:semantic_init}

A key challenge in applying KGE to document-level extraction is handling unseen entities---entities appearing in new documents but absent from the training set. Standard embedding models cannot score triples containing such entities.

\paragraph{Motivation.}
Semantically similar entities tend to occupy analogous structural roles in knowledge graphs—entities of the same type typically participate in similar relation patterns. This suggests that semantic representations from pre-trained language models may provide reasonable proxy for structural embeddings when the latter are unavailable. Moreover, prior work on cross-lingual word embeddings demonstrates that independently trained embedding spaces often exhibit approximate linear alignment~\cite{mikolov2013exploiting}, supporting the feasibility of learning a linear projection between semantic and structural spaces.

\paragraph{Cross-Space Projection.}
Based on this insight, we propose a semantic-guided initialization strategy. For unseen entities, we leverage Chinese-RoBERTa-wwm~\cite{cui2021pretrain}, which encodes entity mentions with rich contextual and type information acquired during pre-training on large-scale Chinese corpora. We project the resulting semantic vectors into RotatE space via a learned linear transformation:
\begin{equation}
\mathbf{e}_{\text{KGE}} = \mathbf{W} \cdot \text{RoBERTa}(e)_{[\text{CLS}]} + \mathbf{b}
\label{eq:projection}
\end{equation}
where $\mathbf{W} \in \mathbb{R}^{1024 \times 768}$ maps the 768-dimensional RoBERTa output to the complex-valued RotatE space (512 dimensions $\times$ 2 for real and imaginary components). The projection is trained on known entities by minimizing the distance between projected and learned embeddings.

This alignment enables scoring any triple regardless of whether its entities appeared during training. For known entities, we use learned RotatE embeddings directly; for unseen entities, we apply the trained projection.

\subsubsection{Tri-partition Validation}
\label{sec:tripartition}

\begin{figure*}[t]
    \centering
    \includegraphics[width=0.8\textwidth]{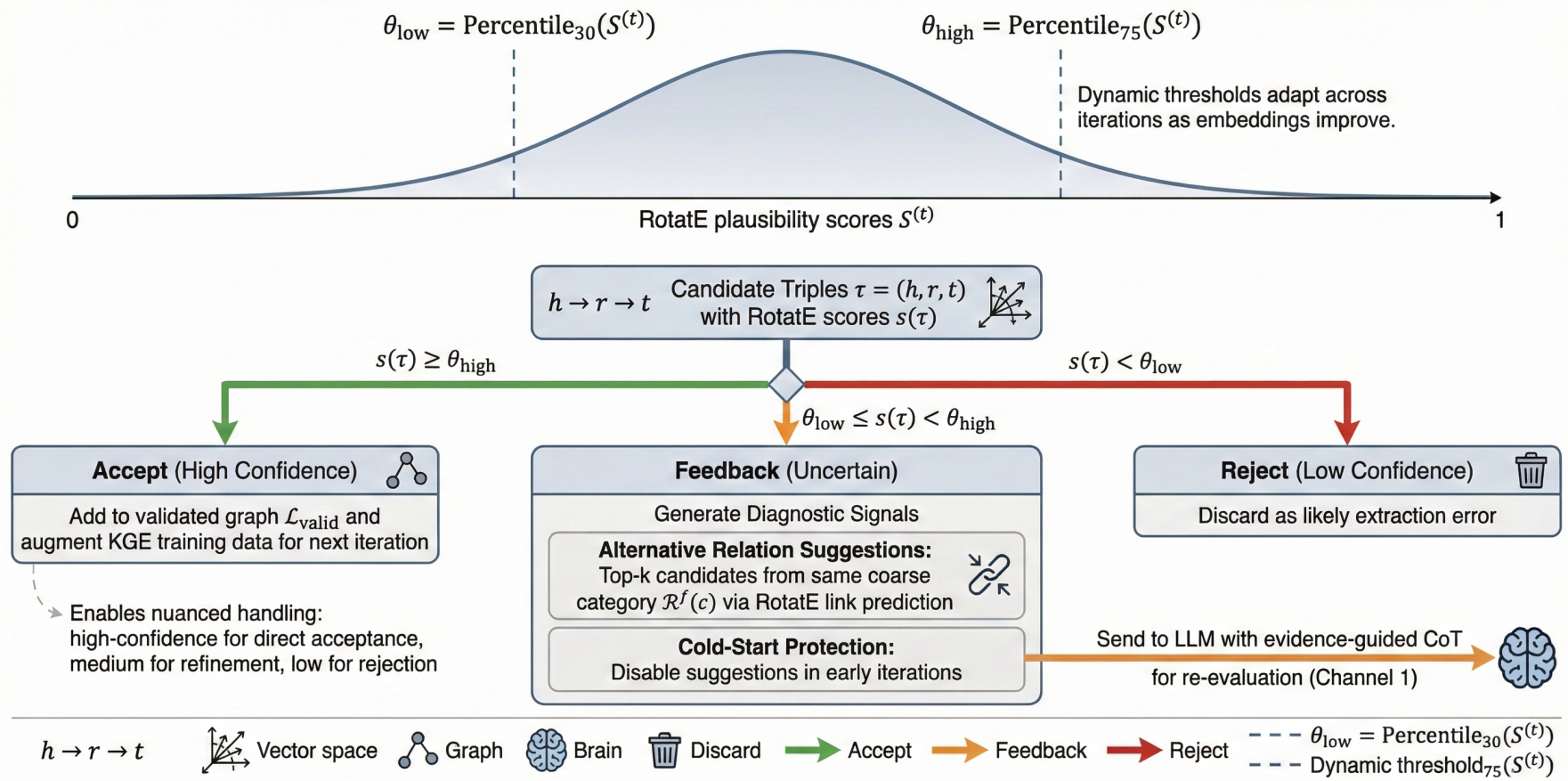}
    \caption{Tri-partition validation mechanism with dynamic thresholds and diagnostic feedback generation.}
    \label{fig:tripartition}
\end{figure*}

Rather than binary accept/reject decisions, we route triples to three paths based on structural scores (Figure~\ref{fig:tripartition}), enabling nuanced handling of uncertain extractions.

\paragraph{Dynamic Thresholds.}
Score distributions shift across iterations as embeddings improve. We use percentile-based thresholds (25th and 70th percentiles) that adapt dynamically to each iteration's score distribution.

\paragraph{Routing Actions.}
Triples are routed based on their scores relative to thresholds:
\begin{equation}
\text{Route}(\tau) = 
\begin{cases}
\textsc{Accept} & \text{if } s(\tau) \geq \theta_{\text{high}} \\
\textsc{Feedback} & \text{if } \theta_{\text{low}} \leq s(\tau) < \theta_{\text{high}} \\
\textsc{Reject} & \text{if } s(\tau) < \theta_{\text{low}}
\end{cases}
\label{eq:tripartition}
\end{equation}
where thresholds are defined as:
\begin{equation}
\theta_{\text{low}}^{(t)} = \text{Percentile}_{25}(\mathcal{S}^{(t)}), \quad \theta_{\text{high}}^{(t)} = \text{Percentile}_{70}(\mathcal{S}^{(t)})
\label{eq:thresholds}
\end{equation}

\textsc{Accept} triples enter the validated graph and augment training data for subsequent iterations. \textsc{Reject} triples are discarded as likely extraction errors. \textsc{Feedback} triples are sent to the LLM with diagnostic signals for re-evaluation (Section~\ref{sec:collaboration}).

\subsubsection{Diagnostic Signal Generation}
\label{sec:diagnostic}

For \textsc{Feedback} triples, we generate actionable diagnostic signals beyond numeric scores to guide LLM re-extraction.

\paragraph{Alternative Relation Suggestion.}
We leverage RotatE's link prediction capability to suggest alternative relations. Given a triple $(h, r, t)$ with low score, we fix entity embeddings and rank candidate relations within the same coarse category $\mathcal{R}^f(c)$, returning the top-$k$ alternatives. This category constraint maintains semantic coherence and prevents drift toward dominant relations.

\paragraph{Cold-Start Protection.}
During early iterations, embeddings lack sufficient training signal. We disable relation suggestions and provide only numeric scores to prevent misleading feedback.

The complete feedback mechanism, including evidence retrieval and LLM re-reasoning, is detailed in Section~\ref{sec:collaboration}.

\subsection{Bidirectional Collaborative Refinement}
\label{sec:collaboration}

\begin{figure*}[t]
    \centering
    \includegraphics[width=0.92\textwidth]{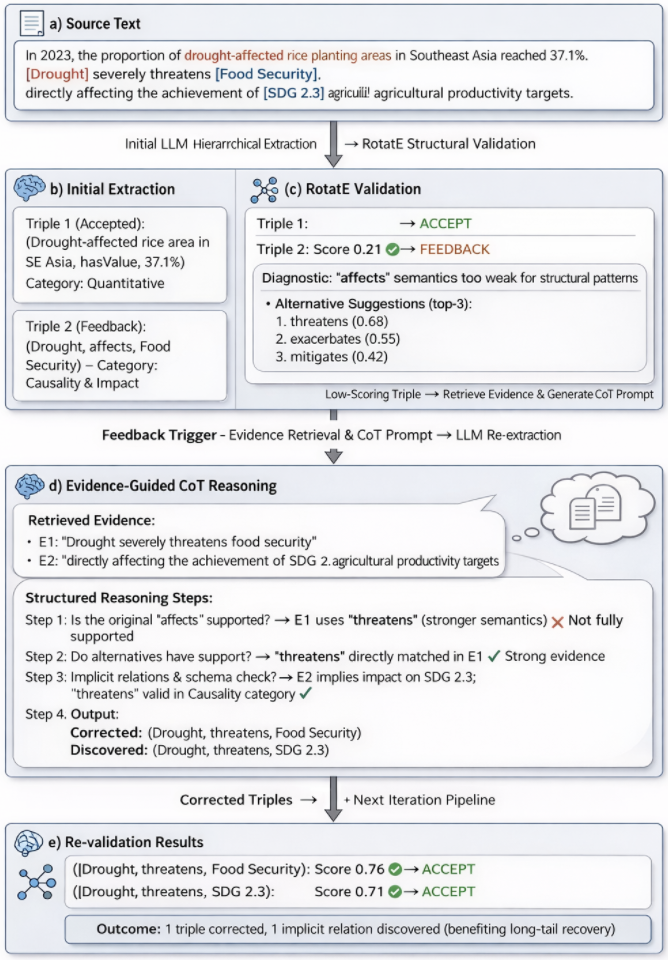}
    \caption{Bidirectional refinement: Channel 1 provides evidence-guided CoT feedback for LLM re-extraction; Channel 2 incorporates validated triples for KGE model updating.}
    \label{fig:collaboration}
\end{figure*}

This section details the two feedback channels that form the core of our iterative refinement mechanism (Figure~\ref{fig:collaboration}).

\subsubsection{Channel 1: Evidence-Guided CoT Feedback (KGE $\to$ LLM)}
\label{sec:channel1}

A limitation of purely score-based feedback is that numeric scores lack textual grounding, potentially leading to hallucination when the LLM cannot verify suggestions against source text. We address this through evidence-guided chain-of-thought reasoning.

\paragraph{Evidence Retrieval.}
For each \textsc{Feedback} triple $(h, r, t)$, we retrieve supporting evidence through entity-anchored sentence retrieval. Given our moderate corpus size, we employ deterministic exact-match: identifying sentences containing both head and tail entity mentions (or their aliases). This ensures high precision—every retrieved sentence is guaranteed to mention the relevant entities. When exact co-occurrence yields insufficient results, we relax to single-entity matching ranked by proximity to relation-indicative keywords.

\paragraph{Chain-of-Thought Prompt.}
We construct a structured prompt (Table~\ref{tab:cot_prompt}) guiding the LLM through explicit reasoning: (1) evaluating whether the original relation is supported by evidence, (2) checking if suggested relations from RotatE have implicit support, (3) verifying schema constraints, and (4) producing corrected triples or confirming rejection.

\begin{table}[t]
\centering
\caption{Evidence-guided CoT feedback prompt template.}
\label{tab:cot_prompt}
\small
\begin{tabular}{p{7.2cm}}
\toprule
\textbf{Evidence-Guided CoT Prompt Template} \\
\midrule
\texttt{[System]} You are an SDG knowledge extraction expert. \\[4pt]
\texttt{[User]} A triple requires re-evaluation. \\[2pt]
\textbf{Original Triple}: (\{head\}, \{relation\}, \{tail\}) \\
\textbf{Structural Score}: \{score\} (threshold: \{threshold\}) \\[2pt]
\textbf{Alternative Relations} (from KGE): \\
\quad 1. \{r1\} \quad 2. \{r2\} \quad 3. \{r3\} \\[2pt]
\textbf{Retrieved Evidence}: \\
\quad E1: ``\{sentence\_1\}'' \\
\quad E2: ``\{sentence\_2\}'' \\[2pt]
\textbf{Instruction}: Reason step by step: \\
1. Is the original relation supported by evidence? \\
2. Do alternative relations have implicit support? \\
3. Are schema constraints satisfied? \\
4. Output: corrected triple as JSON, or ``reject''. \\
\bottomrule
\end{tabular}
\end{table}

\paragraph{Output Processing.}
Corrected triples re-enter the validation pipeline for the next iteration. This mechanism not only corrects extraction errors but also recovers implicit relations beneficial for long-tail types lacking explicit textual markers.

\paragraph{Retry Limit.}
To prevent infinite loops, triples failing validation after $K=3$ attempts are permanently rejected.

\subsubsection{Channel 2: Validated Triple Integration (LLM $\to$ KGE)}
\label{sec:channel2}

Channel 2 progressively improves KGE representations by incorporating validated extractions into the training set.

\paragraph{Confidence-Based Selection.}
Based on RotatE plausibility scores, triples are partitioned into three tiers using percentile thresholds:
\begin{itemize}
    \item \textbf{Top 30\%} (high confidence): Directly added to KGE training set
    \item \textbf{30\%--75\%} (medium confidence): Sent to Channel 1 for LLM verification; added if approved
    \item \textbf{Bottom 25\%} (low confidence): Rejected as likely errors
\end{itemize}

This tiered approach balances precision and recall: high-confidence triples provide reliable training signal, while medium-confidence triples undergo additional verification to recover valid extractions that structural scoring alone might miss.

\paragraph{Incremental KGE Update.}
After each iteration, the KGE model is updated with newly validated triples. We employ warm-start training initialized from the previous iteration's parameters, fine-tuning for a limited number of epochs to incorporate new knowledge while preserving learned patterns.

\subsubsection{Convergence and Termination}
\label{sec:termination}

The iterative process terminates after $T=4$ iterations. Empirically, we observe that the validated triple count stabilizes by iteration 3--4, with diminishing returns from additional cycles. The final knowledge graph comprises all triples accumulated across iterations that passed both structural validation and schema constraints.

\section{Experiments}
\label{sec:experiments}

We conduct comprehensive experiments to evaluate our bidirectional collaborative framework. Our experiments address the following research questions:
\begin{itemize}
    \item \textbf{RQ1}: Does our framework outperform existing extraction methods?
    \item \textbf{RQ2}: How does each component contribute to the overall performance?
    \item \textbf{RQ3}: How effective is the Evidence-Guided CoT Feedback mechanism?
    \item \textbf{RQ4}: How does the iterative refinement process converge?
\end{itemize}

\subsection{Experimental Setup}
\label{sec:setup}

\subsubsection{Dataset Construction}
\label{sec:dataset}

\textbf{Data Source.}
We construct a high-quality, domain-specific SDG knowledge graph dataset from the \textit{Big Earth Data in Support of the Sustainable Development Goals} report series published by the Chinese Academy of Sciences. This authoritative scientific publication exhibits two key characteristics: (1) \textit{Domain Authority}: authored by multidisciplinary scientists covering environmental, economic, and social SDG indicators with standardized terminology; (2) \textit{Textual Complexity}: containing extensive unstructured policy narratives, scientific arguments, and quantitative data, with each report averaging approximately 100,000 Chinese characters.

\textbf{Temporal Split Strategy.}
Unlike conventional random splitting, we adopt a \textit{temporal split} strategy to rigorously evaluate model performance under realistic knowledge evolution scenarios:
\begin{itemize}
    \item \textbf{Training Set (Historical Knowledge)}: Constructed from the 2023 report using an ``LLM-assisted extraction + manual cleaning'' pipeline, yielding 1,548 weakly-supervised triples that simulate learning from historical knowledge.
    \item \textbf{Test Set (Future Knowledge)}: Constructed from the 2024 report through expert-level manual review, producing 1,758 gold-standard triples for rigorous evaluation.
\end{itemize}

This temporal extrapolation setup evaluates the model's ability to generalize to emerging scientific discoveries and policy dynamics, which is more challenging than in-distribution testing and better reflects real-world knowledge graph evolution.

\textbf{Dataset Statistics.}
Table~\ref{tab:dataset} summarizes the dataset characteristics. Despite the low-resource regime, the dataset exhibits high information density with diverse entity and relation types.

\begin{table}[t]
\centering
\caption{Dataset statistics.}
\label{tab:dataset}
\small
\begin{tabular}{lcc}
\toprule
\textbf{Item} & \textbf{Train (2023)} & \textbf{Test (2024)} \\
\midrule
Report Length & $\sim$100K chars & $\sim$100K chars \\
\#Triples & 1,535 & 1,758 \\
\#Relation Types & 74 & 87 \\
\#Unique Entities & 2,132 & 2,761 \\
Avg. Triples per Chunk & 153.50 & 175.80 \\
\midrule
Annotation Method & LLM + Manual & Expert Review \\
\bottomrule
\end{tabular}
\end{table}

\subsubsection{Baselines}
\label{sec:baselines}

We compare our framework against the following baselines:
\begin{itemize}
    \item \textbf{LLM One-shot}: DeepSeek-V3 with a single demonstration example.
    \item \textbf{LLM Few-shot}: DeepSeek-V3 with 5-shot in-context examples covering diverse relation types.
    \item \textbf{OpenIE + Semantic Mapping}: Open information extraction followed by semantic clustering to map extracted relations to the 87 predefined types.
    \item \textbf{Schema-constrained (No Iter.)}: Our extraction pipeline excluding the bidirectional refinement module, serving as a single-pass baseline.
\end{itemize}

\subsubsection{Evaluation Metrics}
\label{sec:metrics}

We adopt Precision, Recall, and F1-score as primary metrics. We report both \textit{Micro-F1} (overall performance) and \textit{Macro-F1} (class-balanced, reflecting long-tail handling). A predicted triple is correct if both the entity pair and relation type match the ground truth. 

To address surface form variations in entity mentions, we employ both exact matching and semantic similarity matching (cosine similarity $\geq$ 0.85 using sentence embeddings). Additionally, we conduct human evaluation on a random sample of 200 triples to assess true extraction quality.

For long-tail analysis, we partition relations into three groups: \textit{Head} ($>$100 samples), \textit{Medium} (20--100 samples), and \textit{Tail} ($<$20 samples).

\subsubsection{Implementation Details}
\label{sec:implementation}

Table~\ref{tab:implementation} summarizes the implementation configurations. All experiments are conducted on NVIDIA A100 GPUs.

\begin{table}[t]
\centering
\caption{Implementation details.}
\label{tab:implementation}
\small
\begin{tabular}{ll}
\toprule
\textbf{Component} & \textbf{Configuration} \\
\midrule
\multicolumn{2}{l}{\textit{LLM Extraction \& Classification}} \\
\quad Model & DeepSeek-V3 (API) \\
\quad Strategy & Hierarchical (8-class → 87-class) \\
\midrule
\multicolumn{2}{l}{\textit{RoBERTa (Frozen)}} \\
\quad Model & Chinese-RoBERTa-wwm-ext \\
\quad Training Data & 2023 + 2024 entities (for alignment) \\
\quad Usage & Semantic Init + MC-Dropout \\
\midrule
\multicolumn{2}{l}{\textit{RotatE Validation}} \\
\quad Embedding Dim & 512 \\
\quad Initial Training & 2023 triples (1,548) \\
\quad Update & Warm-start each iteration \\
\midrule
\multicolumn{2}{l}{\textit{Channel 1: Evidence-Guided CoT}} \\
\quad Evidence Retrieval & Entity-anchored exact match \\
\quad Relation Suggestions & Top-3 from RotatE \\
\midrule
\multicolumn{2}{l}{\textit{Channel 2: Active Selection}} \\
\quad MC-Dropout Runs & 5 \\
\quad Selection & High structure + High uncertainty \\
\midrule
\multicolumn{2}{l}{\textit{Iteration}} \\
\quad Max Iterations & $T=4$ \\
\quad Convergence & $\epsilon=0.01$ \\
\bottomrule
\end{tabular}
\end{table}

\subsection{Main Results (RQ1)}
\label{sec:main_results}

Table~\ref{tab:main_results} compares overall performance. Our framework achieves the highest scores across all metrics.

\begin{table}[t]
\centering
\small
\caption{Overall performance comparison (\%). Best in \textbf{bold}, second-best \underline{underlined}.}
\label{tab:main_results}
\resizebox{\columnwidth}{!}{
    \begin{tabular}{@{}lcccc@{}}
    \toprule
    \textbf{Method} & \textbf{Prec.} & \textbf{Rec.} & \textbf{Micro-F1} & \textbf{Macro-F1} \\
    \midrule
    LLM One-shot & 12.19 & 15.17 & 13.51 & 8.68 \\
    OpenIE + Mapping & 19.85 & 27.47 & 23.62 & 12.58 \\
    LLM Few-shot & \underline{24.16} & \underline{27.08} & \underline{25.54} & \underline{10.19} \\
    \midrule
    \textbf{Ours} & \textbf{34.84} & \textbf{38.96} & \textbf{36.79} & \textbf{21.63} \\
    \bottomrule
    \end{tabular}
}
\end{table}

Our framework yields 36.79\% Micro-F1 and 21.63\% Macro-F1, outperforming the strongest baseline (LLM Few-shot) by 11.25 and 11.44 points, respectively. 

\textbf{Precision.} Our method achieves 34.84\% precision, substantially higher than all baselines. LLM One-shot suffers from severe schema drift and hallucination (12.19\%), while OpenIE produces noisy extractions (19.85\%) due to the difficulty of mapping open-domain relations to predefined types. Our schema-constrained extraction combined with RotatE structural validation effectively filters implausible triples.

\textbf{Recall.} Our framework reaches 38.96\% recall, improving 11.88 points over Few-shot. This gain stems from the Evidence-Guided CoT Feedback (Channel 1), which recovers medium-confidence triples through re-evaluation with textual evidence. Notably, OpenIE achieves moderate recall (27.47\%) by extracting aggressively, but this comes at significant precision cost.

\textbf{F1 Scores.} The substantial Micro-F1 improvement demonstrates that our iterative refinement enhances both precision and recall simultaneously. The Macro-F1 gain (21.63\% vs 10.19\%) is particularly notable, indicating effective long-tail handling through our bidirectional collaboration mechanism. Detailed analysis is provided in Section~\ref{sec:ablation}.

\subsection{Ablation Study (RQ2)}
\label{sec:ablation}

We ablate key components to validate their contributions (Table~\ref{tab:ablation}).

\begin{table}[t]
\centering
\small
\caption{Ablation results (\%).}
\label{tab:ablation}
\begin{tabular}{@{}lccc@{}}
\toprule
\textbf{Variant} & \textbf{Prec.} & \textbf{Rec.} & \textbf{Micro-F1} \\
\midrule
Full Framework & \textbf{34.84} & \textbf{38.96} & \textbf{36.79} \\
w/o Evidence Retrieval & 28.50 & 22.30 & 25.01 \\
w/o Active Selection & 33.20 & 21.80 & 26.35 \\
w/o Semantic Init & 32.00 & 25.40 & 28.30 \\
w/o Iterative Refinement & 24.20 & 19.40 & 21.50 \\
\bottomrule
\end{tabular}
\end{table}

\textbf{Evidence Retrieval (Channel 1).} Removing evidence retrieval causes a substantial drop in both recall (38.96\%$\rightarrow$22.30\%) and Micro-F1 (36.79\%$\rightarrow$25.01\%, -11.78 points). This confirms that structural suggestions alone are insufficient---grounding predictions in source text is critical for recovering implicit relations and correcting extraction errors.

\textbf{Active Selection (Channel 2).} Without uncertainty-based active selection, recall drops significantly to 21.80\%, leading to a Micro-F1 of 26.35\% (-10.44 points). Notably, precision remains relatively stable (33.20\%), indicating that active selection primarily contributes to expanding knowledge coverage by identifying informative samples that the model has not yet learned.

\textbf{Semantic Initialization.} Without RoBERTa-based semantic initialization, Micro-F1 decreases by 8.49 points (36.79\%$\rightarrow$28.30\%). This degradation occurs because RotatE cannot effectively score triples containing out-of-vocabulary entities, confirming that semantic vector initialization is essential for temporal knowledge graph construction.

\textbf{Iterative Refinement.} Removing iteration entirely results in 21.50\% Micro-F1 (-15.29 points), with notable decreases in both precision (24.20\%) and recall (19.40\%). Interestingly, this variant performs below LLM Few-shot (25.54\%), as the initial RotatE model---trained only on historical data---tends to reject valid triples containing unseen patterns. This gap is progressively closed through iterative refinement, validating the cumulative benefit of multi-round collaboration.

\subsubsection{Long-tail Analysis}

Table~\ref{tab:longtail} breaks down F1 by relation frequency.

\begin{table}[t]
\centering
\small
\caption{F1 (\%) by relation frequency.}
\label{tab:longtail}
\begin{tabular}{@{}lccc@{}}
\toprule
\textbf{Method} & \textbf{Head} & \textbf{Medium} & \textbf{Tail} \\
\midrule
LLM Few-shot & 33.1 & 20.8 & 6.7 \\
w/o Iterative & 30.8 & 14.0 & 7.7 \\
\textbf{Full} & \textbf{47.4} & \textbf{28.1} & \textbf{13.3} \\
\bottomrule
\end{tabular}
\end{table}

Our framework consistently outperforms baselines across all frequency groups, with pronounced gains on Tail relations (13.3\% vs 6.7\% for Few-shot, nearly doubling). The iterative mechanism is particularly critical for the long tail, improving Tail F1 from 7.7\% (w/o Iterative) to 13.3\% (+5.6 points). These results validate that bidirectional collaboration effectively addresses the long-tail distribution challenge.

\subsection{Convergence Analysis (RQ4)}
\label{sec:convergence}

Previous experiments evaluated final output quality via F1 scores. This section analyzes the behavioral characteristics of the bidirectional collaboration mechanism across iterations. To enable consistent evaluation across all nine rounds, we adopt precision and validated triple count as convergence metrics, as these can be computed at each iteration without requiring full re-evaluation on the test set. We extend the iteration rounds to 9 to comprehensively observe convergence behavior and potential overfitting. Table~\ref{tab:convergence} and Figure~\ref{fig:convergence} present the detailed metrics and convergence curves.

\begin{table}[t]
\centering
\caption{Validation metrics across iteration rounds.}
\label{tab:convergence}
\small
\begin{tabular}{ccc}
\toprule
\textbf{Round} & \textbf{\#Validated Triples} & \textbf{Precision (\%)} \\
\midrule
1 & 1100 & 29.1 \\
2 & 1300 & 31.5 \\
3 & 1530 & 34.0 \\
4 & 1965 & 34.9 \\
5 & 2060 & \textbf{36.7} \\
6 & 1905 & 35.7 \\
7 & 1700 & 33.5 \\
8 & 1518 & 30.6 \\
9 & 1400 & 28.0 \\
\bottomrule
\end{tabular}
\end{table}

The iterative process exhibits three distinct phases:

\vspace{1em}
\noindent
\begin{minipage}{\columnwidth}
    \centering
    \includegraphics[width=0.95\linewidth]{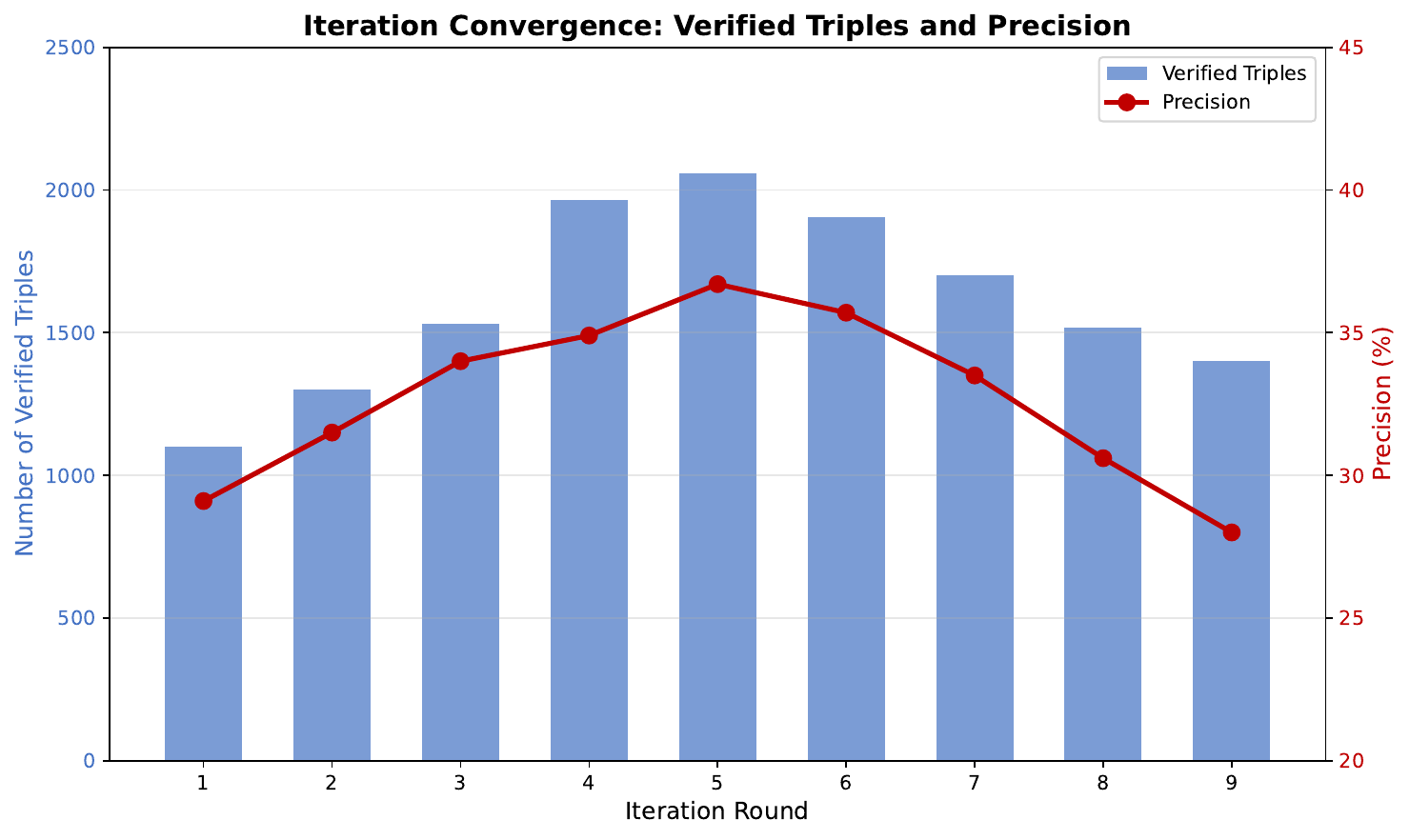}
    \captionof{figure}{Convergence over iterations: validated triple count and precision (\%).}
    \label{fig:convergence}
\end{minipage}
\vspace{1em}

\textbf{Collaborative Growth (Round 1--4).} During early iterations, the bidirectional collaboration mechanism progressively takes effect. The RotatE validator strengthens its structural scoring capability with increasing training data, while the CoT evidence feedback recovers correctly extracted triples that were conservatively rejected in earlier rounds. The synergy drives validated triples from 1,100 to 1,965 and precision from 29.1\% to 34.9\%, demonstrating simultaneous improvement in both quantity and quality. Notably, the Round 4 precision (34.9\%) closely aligns with the overall framework precision (34.84\%) reported in Section~\ref{sec:main_results}, confirming the consistency of our evaluation.

\textbf{Peak Performance (Round 5--6).} The framework reaches optimal performance at Round 5, with 2,060 validated triples and 36.7\% precision. At this stage, RotatE and CoT feedback achieve their best collaboration: RotatE accurately identifies structurally plausible triples while CoT provides semantic-level verification for borderline cases. Round 6 shows a slight decline (1,905 triples, 35.7\% precision) but remains at a high level, indicating a brief plateau before degradation.

\textbf{Overfitting Decay (Round 7--9).} From Round 7 onward, performance degrades continuously. Validated triples decrease from 1,700 to 1,400, and precision drops from 33.5\% to 28.0\%. We hypothesize this degradation is primarily driven by RotatE overfitting: after excessive training iterations, the model becomes overly sensitive to specific patterns in historical data, leading to distorted scoring criteria that both reject valid borderline triples and memorize noise patterns. Notably, Round 9 precision (28.0\%) falls below Round 1 (29.1\%), indicating that over-iteration not only fails to yield gains but erases earlier improvements.

\textbf{Validation of Early Stopping.} These results validate the necessity of early stopping. While Round 5 achieves peak precision (36.7\%), the declining trend from Round 6 suggests that stopping at Round 4 ($T=4$) offers better stability margins. The Round 4 configuration balances two advantages: its precision (34.9\%) is already close to the peak with limited loss, and it provides a sufficient safety buffer against overfitting. The final performance reported in Section~\ref{sec:main_results} (Micro-F1 = 36.79\%) is based on $T=4$, which achieves a favorable trade-off between extraction quality and system robustness.

\textbf{Key Insights.} (1) Bidirectional collaboration achieves simultaneous quantity-quality improvement in early iterations, validating the synergy between structural validation and evidence-guided feedback. (2) A clear performance peak exists (Round 5), beyond which overfitting causes continuous degradation. (3) Over-iteration can regress performance below initial levels, making early stopping ($T=4$) essential for stable output.

\section{Discussion}
\label{sec:discussion}

\subsection{Why Does Bidirectional Collaboration Work?}
\label{sec:why_work}

The effectiveness of our framework stems from the complementary strengths of LLMs and knowledge graph embeddings.

\paragraph{LLMs: Strong Semantics, Weak Structure.}
Large language models excel at semantic understanding and contextual reasoning, enabling accurate extraction of entity mentions and relation expressions. However, LLMs operate on local context and lack awareness of global graph structure, leading to two failure modes: (1) extracting locally plausible but globally inconsistent triples, and (2) hallucinating relations without structural support.

\paragraph{Embeddings: Strong Structure, Weak Semantics.}
Knowledge graph embeddings like RotatE capture global structural patterns through geometric relationships in latent space. They can identify implausible triples by detecting structural anomalies, but cannot interpret natural language or extract new knowledge from text.

\paragraph{Bidirectional Synergy.}
Our framework creates a virtuous cycle through two complementary channels:
\begin{itemize}
    \item \textbf{Channel 1} (KGE $\to$ LLM): The evidence-guided Chain-of-Thought feedback mechanism transforms structural signals into actionable guidance. Rather than merely rejecting low-scoring triples, it retrieves supporting evidence from source text and suggests structurally plausible alternatives, enabling structure-aware reasoning grounded in textual evidence.
    \item \textbf{Channel 2} (LLM $\to$ KGE): Validated high-confidence triples progressively enrich the embedding training set, improving structural representations across iterations and expanding coverage of domain-specific patterns.
\end{itemize}

The ablation study (Section~\ref{sec:ablation}) validates this synergy: removing either channel causes substantial performance degradation, confirming that neither component alone achieves the full framework's effectiveness.

\subsection{Limitations}
\label{sec:limitations}

\paragraph{Entity Coverage in Evolving Corpora.}
Our semantic initialization addresses the out-of-vocabulary problem by projecting unseen entities into the embedding space via a learned alignment layer. However, this approach assumes that language model representations transfer effectively to new entities. When entity surface forms differ substantially from training data, the projection quality may degrade.

\paragraph{Cold Start Instability.}
During early iterations, RotatE is trained on limited data, leading to less reliable structural suggestions. Our framework mitigates this through conservative initial thresholds and cold-start protection that disables relation suggestions until sufficient training signal accumulates.

\paragraph{Domain Specificity.}
Our evaluation focuses on Chinese SDG reports with a manually designed 89-relation schema. While the framework architecture is domain-agnostic, adaptation to other domains requires schema redesign. The effectiveness may also vary across languages due to differences in pre-trained language model coverage.

\paragraph{Computational Overhead.}
The iterative refinement process requires multiple rounds of LLM inference and embedding model updates. For resource-constrained settings, a trade-off between iteration depth and computational cost must be considered.

\subsection{Future Work}
\label{sec:future_work}

\paragraph{End-to-End Alignment Learning.}
Currently, the projection layer for semantic initialization is trained separately. Future work could explore joint optimization that updates both the projection and KGE embeddings simultaneously, potentially improving cross-space alignment.

\paragraph{Improved Evidence Retrieval.}
Our evidence retrieval relies on entity-anchored exact matching for high precision. Incorporating dense retrieval methods could improve coverage for paraphrased mentions while maintaining grounding quality.

\paragraph{Automated Schema Induction.}
The current schema was manually designed for SDG documents. Investigating automated schema induction through clustering or LLM-guided ontology generation would reduce manual effort and enable rapid adaptation to new domains.

\paragraph{Cross-domain and Multilingual Extension.}
Evaluating LEC-KG on diverse domains (e.g., climate policy, public health) and languages would establish the generalizability of our bidirectional collaboration paradigm.


\section{Conclusion}
\label{sec:conclusion}

We presented LEC-KG, a bidirectional collaborative framework that integrates the semantic understanding of Large Language Models with the structural reasoning of Knowledge Graph Embeddings for domain-specific knowledge graph construction. Our approach establishes mutual enhancement between the two paradigms: KGE provides structure-aware feedback to refine LLM extractions through evidence-guided Chain-of-Thought reasoning, while validated triples progressively improve KGE representations. To address practical challenges, we introduced semantic initialization for handling unseen entities and hierarchical coarse-to-fine classification for mitigating long-tail relation bias. Experiments on Chinese Sustainable Development Goal reports demonstrate substantial improvements over LLM baselines, with notable gains on low-frequency relations. Through iterative refinement, LEC-KG reliably transforms unstructured policy text into validated knowledge graph triples, offering a practical solution for domain-specific knowledge graph construction.

\printcredits

\appendix

\section{Complete Hierarchical Relation Schema}
\label{app:relation_schema}

Table~\ref{tab:relation_schema_full} presents the complete hierarchical relation schema with all 89 fine-grained relation types. Relations marked with $\dagger$ are long-tail types with frequency $< 5$ in our annotated dataset.

\begin{table*}[t]
\centering
\caption{Complete hierarchical relation schema. Relations marked with $\dagger$ are long-tail types (frequency $< 5$).}
\label{tab:relation_schema_full}
\small
\begin{tabularx}{\textwidth}{p{2.8cm} X c}
\toprule
\textbf{Category} & \textbf{Fine-grained Relations} & \textbf{Count} \\
\midrule
\textit{Definition \& Naming} & definedAs, fullNameOf, abbreviationOf, aliasOf$^\dagger$ & 4 \\
\midrule
\textit{Hierarchy \& Composition} & belongsTo, contains, composedOf, partOf & 4 \\
\midrule
\textit{Spatiotemporal} & locatedIn, distributedIn, covers, timePointOf, timeRangeOf, startsAt, endsAt, publishedOn, implementedAt, spatialDistributionPattern, displayScale & 11 \\
\midrule
\textit{Quantitative} & hasValue, hasUnit, valueRangeOf, maxValueOf, minValueOf, thresholdOf, precisionOf, meanValueOf, medianOf, stdDevOf, spatialResolution, temporalResolution, gridSizeOf, updateFrequency & 14 \\
\midrule
\textit{Trend \& Change} & trendOf, changeAmountOf, changeRateOf, yoyChangeOf, momChangeOf, growthRateOf$^\dagger$ & 6 \\
\midrule
\textit{Provenance \& Method} & dataSourceOf, publishedBy, providedBy, producedBy, citedFrom, hasInput$^\dagger$, hasOutput$^\dagger$, usesMethod, usesModel, usesAlgorithm$^\dagger$, integratedWith, basedOnData, basedOnAssumption$^\dagger$, evaluationMethod, validationMethod, monitoringIndicator, monitoringDataType$^\dagger$ & 17 \\
\midrule
\textit{Causality \& Impact} & causes, affects, threatens, promotes, reduces, mitigates, exacerbates, dependsOn, constrainedBy, requires & 10 \\
\midrule
\textit{Application \& Status} & usedFor, appliedTo, supportsDecision, constructs, deploys, integrates, implements, hasFunction, canIdentify, canDetect$^\dagger$, statusOf, progressOf$^\dagger$, completionOf, milestoneOf, versionOf, hasProblem, hasDefect, bottleneckOf, riskOf, relatedToSDG, correspondsToSDG$^\dagger$, contributesToSDG, promotesSDG & 23 \\
\midrule
\multicolumn{2}{r}{\textbf{Total}} & \textbf{89} \\
\bottomrule
\end{tabularx}
\end{table*}
\bibliographystyle{unsrtnat}

\bibliography{references}

\end{document}